%% file: main.tex
\documentclass[10pt,twocolumn,letterpaper]{article}

\usepackage{cvpr}              
\usepackage{xcolor}         
\usepackage{graphicx}

\input{preamble}

\definecolor{cvprblue}{rgb}{0.21,0.49,0.74}
\usepackage[pagebackref,breaklinks,colorlinks,allcolors=cvprblue]{hyperref}

\def\paperID{5822} 
\def\confName{CVPR}
\def\confYear{2025}

\newcommand{\acronym}{SimAvatar\xspace}
\title{\textit{\acronym}: Simulation-Ready Avatars with Layered Hair and Clothing}

\author{Xueting Li,\, Ye Yuan, Shalini De Mello,\, Gilles Daviet, Jonathan Leaf,\, Miles Macklin,\\ Jan Kautz,\, Umar Iqbal\\
\\
NVIDIA\\
\\
\url{https://nvlabs.github.io/SimAvatar}
}

\begin{document}
\twocolumn[{%
\renewcommand\twocolumn[1][]{#1}%
\maketitle
\begin{center}
\centering
\vspace{-5mm}
\includegraphics[trim={0 0 2.5cm 0},clip, width=\linewidth]{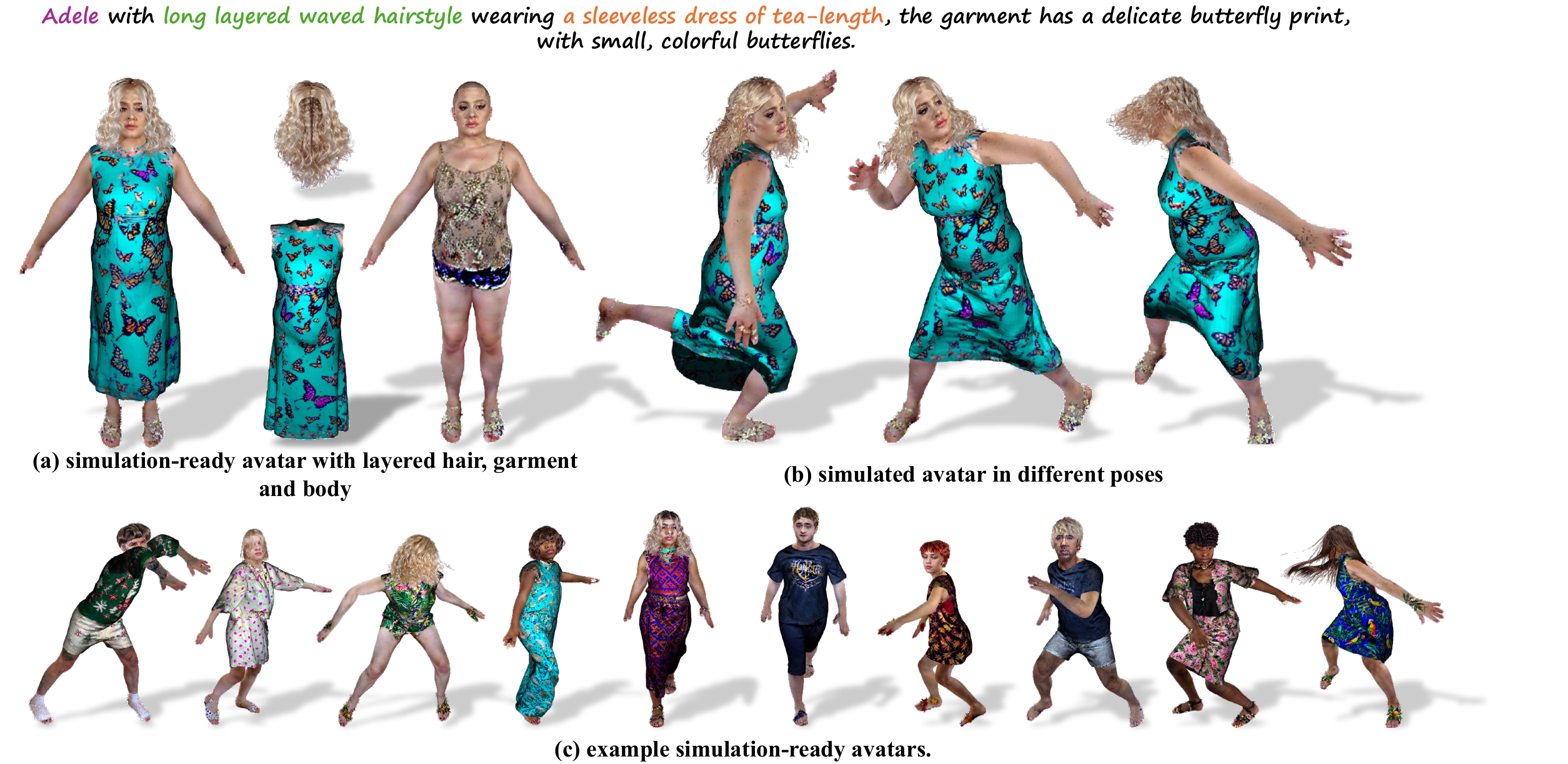}
\captionof{figure}{(a) \acronym synthesizes  simulation-ready 3D avatars with layered hair, body and clothing. (b) By leveraging physics-based simulators, \acronym produces realistic pose-dependent motion effects such as flowing hair strands and garment wrinkles. (c) \acronym produces simulation-ready 3D avatars with diverse identities, garments, and hairstyles.}
\label{fig:teaser}
\end{center}%
}]

\input{sec/0.Abstract}
\vspace{-5mm}
\input{sec/1.Introduction}
\input{sec/2.RelatedWork}

\input{sec/3.Method}
\input{sec/4.Experiments}
\input{sec/5.Conclusions}

{
    \small
    \bibliographystyle{ieeenat_fullname}
    \bibliography{main}
}

\clearpage
\input{supplementary}

\end{document}

%% file: preamble.tex
\newcommand{\norm}[1]{\left\lVert#1\right\rVert}


%% file: sec/0.Abstract.tex
\begin{abstract}
We introduce SimAvatar, a framework designed to generate simulation-ready clothed 3D human avatars from a text prompt. Current text-driven human avatar generation methods either model hair, clothing, and the human body using a unified geometry or produce hair and garments that are not easily adaptable for simulation within existing simulation pipelines. The primary challenge lies in representing the hair and garment geometry in a way that allows leveraging established prior knowledge from foundational image diffusion models (e.g., Stable Diffusion) while being simulation-ready using either physics or neural simulators. To address this task, we propose a two-stage framework that combines the flexibility of 3D Gaussians with simulation-ready hair strands and garment meshes. Specifically, we first employ three text-conditioned 3D generative models to generate garment mesh, body shape and hair strands from the given text prompt. To leverage prior knowledge from foundational diffusion models, we attach 3D Gaussians to the body mesh, garment mesh, as well as hair strands and learn the avatar appearance through optimization. To drive the avatar given a pose sequence, we first apply physics simulators onto the garment meshes and hair strands. We then transfer the motion onto 3D Gaussians through carefully designed mechanisms for each body part. As a result, our synthesized avatars have vivid texture and realistic dynamic motion. To the best of our knowledge, our method is the first to produce highly realistic, fully simulation-ready 3D avatars, surpassing the capabilities of current approaches.
\end{abstract}

%% file: sec/1.Introduction.tex
\section{Introduction}
\vspace{-2mm}


The creation of photo-realistic and dynamic human avatars has a wide range of applications, including virtual try-on, film and gaming production, virtual assistants, AR/VR, and telepresence. Traditionally, this process has required specialized training, making it inaccessible to general users. Recently, advancements in foundational diffusion models have accelerated research efforts aimed at democratizing 3D human avatar creation, enabling easy user control through text~\cite{cao2023dreamavatar, kolotouros2023dreamhuman, yuan2024gavatar, liao2024tada} or images~\cite{huang2024tech}.

Earlier approaches for 3D human avatar creation represent the hair, body, and clothing as a single layer, making it difficult to independently simulate or edit each region due to their entangled geometry. To address this limitation, recent works have used layered structures to separately represent the body, garments, or hair~\cite{wang2023disentangled, HumanLiff, dong2024tela, zhang2023teca}. However, many of these approaches rely on implicit representations, such as NeRF~\cite{mildenhall2020nerf}, to define garment or hair geometry. Although implicit representations facilitate leveraging prior knowledge from foundational diffusion models, they are challenging to animate within existing simulators, limiting the realistic motion of hair and garments due to body movements. As a result, these methods struggle to produce avatars that look realistic when animated.


Thus, a natural question arises: can we design a 3D avatar generation pipeline that can leverage rich prior knowledge from image diffusion models, while being readily compatible with existing simulation pipelines? The key challenge in addressing this question lies in connecting the different representations used in current simulators and text-driven avatar generation pipelines. The former typically requires a smooth and clean non-watertight mesh or specifically designed hair strands, whose topologies are inflexible for optimization and hard to constrain. The latter often adopt implicit representations (such as NeRF~\cite{mildenhall2020nerf} or SDF~\cite{wang2021neus}) which, while optimizable by noisy supervision signals from diffusion models, cannot be easily converted to an open mesh or hair strands suitable for simulation.

To address these issues, we present a novel framework, \acronym, which generates a 3D human avatar from a text prompt that can be readily animated by existing hair and garment simulators.
The key idea is to adopt suitable representations for different human parts (e.g., hair, body, and garments) and leverage the prior knowledge of both image diffusion models and simulators.
To this end, we propose representing the geometry of human hair, body, and garment using hair strands, parametric body model SMPL~\cite{SMPL:2015}, and meshes, respectively. 
While these representations are simulation-ready, they pose challenges to learning realistic texture details. 
To resolve this issue, we further attach 3D Gaussians to these coarse geometric representations. 3D Gaussians are not only flexible and can be easily driven by the mesh or hair strands, but they also show impressive capacity to model realistic appearance when combined with image diffusion models~\cite{yuan2024gavatar,liu2023humangaussian}.

Given a text prompt, we first generate hair strands, body mesh, and garment mesh using three separate text-conditioned 3D generative models trained specifically for each body region.
We then attach and learn 3D Gaussians through Score Distillation Sampling (SDS) based optimization using a powerful diffusion model trained on human images~\cite{rv5.1}. 
To drive the avatar using a given pose sequence, we animate the body using linear blend skinning while the hair strands and garment mesh are simulated using their respective simulators~\cite{admm2023gilles, grigorev2023hood}. Next, we transfer the motion from the strands or meshes onto 3D Gaussians through a carefully designed mechanism to ensure they conform to the mesh or strand motion. 
During inference, the avatar can be easily animated by novel pose sequences. Our novel design yields avatars with both realistic appearance and dynamic motion with pose-dependent deformations of the hairs and garment as shown in Fig.~\ref{fig:teaser}. 
Our contributions are summarized as follows:
\begin{itemize}[leftmargin=15pt]
    \item To the best of our knowledge, this is the first work that generates fully simulation-ready 3D avatars with separate layers for the body, garment, and hair.
    \item We propose a text-conditioned diffusion model to generate plausible garment geometry given text inputs and show how existing text-conditioned hair strands and body shape generators can be efficiently leveraged. 
    \item We use 3D Gaussians on top of the body mesh, hair strands, and garment mesh to synthesize high-fidelity appearance details. We optimize the properties of  Gaussians using carefully designed SDS-based optimization~\cite{xu2022dream3d} and ensure plausible texture disentanglement using additional regularization and prompt engineering. 
    \item Our approach produces avatars with detailed geometry, realistic texture, as well as dynamic garment and hair motion, resulting in significantly better quality than existing methods. 
\end{itemize}

%% file: sec/2.RelatedWork.tex
\section{Related Work}
\vspace{-2mm}
\label{sec:related_work}

\subsection{Text to 3D Avatar Generation}
\vspace{-1mm}
The domain of text-to-3D generation has witnessed significant advancements ~\cite{poole2022dreamfusion, lin2022magic3d, richardson2023texture, chen2023fantasia3d, wang2023prolificdreamer, tang2023dreamgaussian} driven by the advent of large text-to-image models~\cite{rombach2022high, saharia2022photorealistic, dalle2, zhang2023adding}. Following the success in generating static 3D objects, various methods have also been developed to create animatable human avatars~\cite{clipmatrix, cao2023dreamavatar, kolotouros2023dreamhuman, jiang2023avatarcraft, zhang2023getavatar, huang2023dreamwaltz, zhang2023avatarverse, huang2024humannorm, huang2024tech, liao2024tada}. ClipMatrix~\cite{clipmatrix} was among the first to create animatable avatars based on textual descriptions, using a CLIP-embedding loss to optimize a mesh-based representation. AvatarClip~\cite{hong2022avatarclip} followed a similar approach but with a NeRF-based representation~\cite{wang2021neus}.  DreamAvatar~\cite{cao2023dreamavatar} and AvatarCraft~\cite{jiang2023avatarcraft} used SDS loss instead of CLIP and learned the NeRF representation in canonical space by integrating human body priors from SMPL~\cite{SMPL:2015}. DreamHumans~\cite{kolotouros2023dreamhuman} introduced a pose-conditioned NeRF representation to also model pose-dependent deformations. DreamWaltz~\cite{huang2023dreamwaltz}, AvatarVerse~\cite{zhang2023avatarverse} and HumanNorm~\cite{zhang2023adding} leverage ControlNets~\cite{zhang2023adding} and demonstrate improved avatar quality with conditional SDS.  AvatarBooth~\cite{Zeng2023AvatarBoothHA} fine-tunes region-specific diffusion models, highlighting that using dedicated models for distinct body regions improves avatar quality. Other methods explore different representations to model the avatars. TADA~\cite{liao2024tada} demonstrates that a mesh-based approach with adaptive mesh subdivision can create high-quality avatars. More recently, GAvatar~\cite{yuan2024gavatar} adopted a primitive-based 3D Gaussian representation, allowing faster rendering in higher resolution and showing higher quality. All of the aforementioned methods, however, represent the human body, hairs, and
garment geometry using a single layer and animate the avatar using linear blend skinning. Hence, they cannot model physically plausible and pose-dependent geometric deformations of the avatar's clothing and hair resulting in avatars that do not appear realistic when animated. 

There are only a few works that generate avatars with separate body and garment or hair layers. SO-SMPL~\cite{wang2023disentangled} uses SMPL+D~\cite{pons2017clothcap} to separately represent the cloth layers, but cannot capture loose clothing due to the inherent limitations of SMPL+D representation. HumanLiff~\cite{HumanLiff}, HumanCoser~\cite{wang2023humancoser}, and TELA~\cite{dong2024tela} use NeRF-based representations and progressively generate minimal-clothed human body and layer-wise clothes. GALA~\cite{kim2024gala} takes an existing single-layer avatar and disentangles it into separate layers using DMTet representation~\cite{gao2020learning} and SDS~\cite{poole2022dreamfusion} optimization. TECA~\cite{zhang2023teca} aims to model hairs in a separate layer but adopts a NeRF-based representation. However, the meshes extracted from NeRF or DMTet tend to be very noisy and cannot be simulated easily using cloth or hair simulators. Our approach, on the other hand, generates avatars with simulation-ready garments and hair. It can model loose clothing, realistic hair with varied lengths and hairstyles, realistic texture, and dynamic motion. 


\subsection{Garment Modeling and Simulation} 
\vspace{-1mm}
One of the most common methods for modeling garments is SMPL+D~\cite{pons2017clothcap,zhang2017detailed,alldieck2018video,ThiemoAlldieck2019Tex2ShapeDF,alldieck2019learning,bhatnagar2019mgn,ma2020learning,bhatnagar2020loopreg}, which adds vertex displacements to the SMPL body mesh to capture the geometry of clothed human bodies. However, this approach struggles to represent loose clothing and dresses due to its fixed topology. To address this limitation, many methods propose modeling clothing separately. 
DiffAvatar~\cite{li2024diffavatar} presents a method for reconstructing untextured garment meshes from scan images using a differentiable simulator.
Garment3DGen~\cite{sarafianos2024garment3dgen} reconstructs textured garment meshes from a single input image by optimizing and deforming a base garment template to align with the target pose.
BCNet~\cite{jiang2020bcnet} directly predicts garment type and shape parameters using a learned regressor, while DeepCloth~\cite{deepcloth_su2022} represents garments through a canonical UV-representation, encoding them in a shared latent space. More recent methods utilize implicit unsigned distance fields for their flexibility in modeling arbitrary topologies and handling open surfaces, which are crucial for cloth simulators~\cite{corona2021smplicit,ren2022dig,guillard2022udf,de2023drapenet,Chen2024NeuralABC}. However, these methods typically focus solely on modeling clothing geometry, neglecting the underlying human body, hair, and texture details. In this work, we build on these methods and propose a novel text-conditioned diffusion model to generate garment meshes using text prompts, while also generating avatars with complete texture and appearance details for garment, hair, and body layers. 


For body motion-driven garment simulation, traditional methods use physics simulators to drape garments on the human body while modeling their material properties~\cite{choi2005stable, Volino09ASimple, bhat2003estimating, Wang11DataDriven}, and avoiding collisions with the underlying body~\cite{baraff1998large, Thomaszewski2008AsynchronousCS}. Differentiable physics simulators have also been proposed to integrate physics into learning and optimization-based frameworks~\cite{Liang19Differentiable, Li22DiffCloth}. While these methods produce highly realistic garment animations, physics simulation is generally computationally expensive and can become unstable in cases of severe interpenetration between the garment and body layers.
To overcome these limitations, more recent works propose neural simulators, which are neural networks trained to generate garment deformations based on body pose and shape information~\cite{guan2012drape, santesteban2019learning, ma2020learning, patel2020tailornet, santesteban2021garmentcollisions, santesteban2022snug, grigorev2023hood}. Earlier methods were trained in a garment-specific manner and could not generalize across different garment types~\cite{Santesteban_2021_CVPR, pan2022predicting}. However, more recent methods can handle various garment topologies, including loose clothing and dresses~\cite{bertiche2020cloth3d, zakharkin2021point, bertiche2021deepsd, santesteban2021ulnefs, bertiche2020pbns, santesteban2022snug, grigorev2023hood}.
While our method can utilize both physics-based and neural simulators, here we use HOOD~\cite{grigorev2023hood} as the neural simulator. 

\subsection{Hair Modeling and Simulation}
\vspace{-1mm}
For hair modeling, strand-based representations are widely favored due to their compatibility with physics simulators and ease of geometric manipulation~\cite{Yuksel2009HairM, piuze2011generalized, shen2023CT2Hair, fei2017multi, hsu2023sagfree, admm2023gilles}. The main challenge in image- or text-guided generation of hair strands lies in capturing the hair's complex appearance, especially the inner strands that are often occluded. For geometry, recent approaches use data-driven priors based on diffusion models to estimate plausible inner strands despite their occlusion in observed data~\cite{sklyarova2023neural, sklyarova2024haar}. For appearance, existing methods employ a strand-aligned 3D Gaussian representation to effectively model intricate hair textures and varied strand thickness~\cite{luo2024gaussianhair, zakharov2024gh}. In this work, we build upon these methods for full-body avatar generation. Specifically, we use a diffusion-based text-to-hair generation model~\cite{sklyarova2024haar} to create hair strands from text prompts. We then attach 3D Gaussians~\cite{zakharov2024gh} to the generated strands to model hair appearance and refine hairstyle according to the given text prompt. Finally, we incorporate a hair simulator~\cite{admm2023gilles} during avatar creation to simulate hairs for varied poses which allows for optimizing the appearance of the inner strands and ensures proper disentanglement of face and hair textures.

%% file: sec/3.Method.tex
\input{figs/overview}

\section{Method}
\vspace{-2mm}
Given a text prompt $\mathcal{T}$, we introduce a novel framework, \acronym, for synthesizing simulation-ready 3D avatars with layered representations of the body, hair, and garment. Unlike traditional text-to-avatar methods that use a single geometry for the entire avatar~\cite{yuan2024gavatar, liao2024tada, chen2023fantasia3d}, our approach decomposes the avatar geometry into distinct layers, each leverages a representation tailored to its simulation and geometry modeling. 
This layered structure not only enhances realism but also enables robust simulation of each component, dramatically improving flexibility in animation and customization. 
Specifically, our avatar geometry is represented by three distinct layers: (1) \textbf{Body mesh}, defined by the SMPL parametric model~\cite{SMPL:2015}, providing a robust foundation for realistic body shapes and movements; (2) \textbf{Garment mesh}, generated using a text-conditioned diffusion model, which produces smooth clothing meshes that are ready for simulation (see Sec.~\ref{sec:diffusion}); (3) \textbf{Hair strands}, modeled using a text-conditioned diffusion model, allowing for the generation of complex hairstyles with strand-level detail (see Sec.~\ref{sec:preliminaries}). 
This approach allows our method to readily handle dynamic scenarios. Given a target body pose sequence $\mathcal{P} = \{p_1, ..., p_n\}$, the body mesh is deformed through linear blend skinning~\cite{SMPL:2015} while both the garment mesh and hair strands are simulated using advanced physics-based simulators (Sec.~\ref{sec:preliminaries}), ensuring realistic motion under various conditions.

To further enable our layered representation to model detailed avatar appearance, including body, hair, and garment, we propose a 3D Gaussian Splatting (3DGS)-based appearance model that builds on top of the layered representation. In particular, we attach 3D Gaussians to the layered geometry of body, garment, and hair, while customizing the Gaussians for each body part to capture the unique structural and dynamic properties of the part. 
These 3D Gaussians are optimized using Score Distillation Sampling (SDS)~\cite{poole2022dreamfusion, yuan2024gavatar}, allowing us to refine appearance details in a way that leverages powerful diffusion priors (Sec.~\ref{sec:gaussian_avatar}). 
As a result, the generated avatars not only look more realistic but also maintain high fidelity during animation. 
An overview of our framework is shown in Fig.~\ref{fig:overview}.


\subsection{Preliminaries}
\vspace{-1mm}
\label{sec:preliminaries}
\paragraph{Garment simulation.}
In this work, we utilize a neural simulator (i.e., HOOD~\cite{grigorev2023hood}) to produce a simulated garment mesh sequence. 
Given a garment mesh $g_0$ and the target body pose sequence $\mathcal{P}$, HOOD first computes the SMPL body mesh corresponding to the SMPL parameters. 
By treating the body meshes as obstacles, HOOD utilizes a GNN to predict the physical status (e.g., position, velocity, etc.) of each vertex on the garment mesh and produce a simulated garment sequence denoted as $\mathcal{G}=\{g_1,...,g_n\}$. More details are discussed in~\cite{grigorev2023hood}.

\paragraph{Strand-based hair representation.}
%
%
%
In this work, we choose strand-based hair representation to capture the thin and layer-wise structure of hairs. 
%
We represent hair as a point cloud $h_0\in \mathrm{R}^{{N_s}\times{N_l}\times 3}$ that is composed of $N_s$ hair strands, with each strand includes $N_l$ line segments.

\paragraph{Hair simulation.}
Given hair strands $h_0$, a target body mesh sequence $\mathcal{P}$ and the simulated garment sequence $\mathcal{G}$, we treat both the body and garment in each time step as obstacles and leverage a hair simulator~\cite{bergou2008discrete,bergou2010discrete,daviet2020simple,daviet2023interactive} to produce the animated hair strand sequence denoted as $\mathcal{H}=\{h_1, ..., h_n\}$. This sequence provides a strong prior that drives the 3D Gaussians attached to the strands to produce high-fidelity hair motion, as will be discussed in Sec.~\ref{sec:gaussian_avatar}.

\subsection{Text-based garment diffusion model}
\vspace{-1mm}
\label{sec:diffusion}
\input{figs/vae}

To ensure compatibility with existing simulators, in this work, we choose to use meshes as the coarse garment geometry representation.
A straightforward approach to modify the garment mesh to match the given text prompt is through SDS-based optimization~\cite{wang2023disentangled,liao2024tada}.
However, optimizing the non-watertight garment mesh in a zero-shot manner is a non-trivial task. On one hand, the topology of garment mesh is hard to change with noisy supervision from foundational diffusion models.
On the other hand, the garment simulators all require a clean, compact, and smooth garment mesh as input.
To resolve this issue, inspired by recent successes in 3D generative models~\cite{zhang20233dshape2vecset,zhang2024clay,wu2024direct3d}, we utilize large-scale garment datasets~\cite{KorostelevaGarmentData,bertiche2020cloth3d} and learn a text-based diffusion model for garment mesh generation.
Specifically, we first train a Variational Auto-encoder (VAE) to learn a compact latent space capturing garment geometry distribution. We then learn a conditional latent diffusion model within the latent space to generate garment meshes from text prompts.
%

\paragraph{VAE for garment mesh reconstruction.}
We adopt the vectorized latent space representation~\cite{zhang20233dshape2vecset,zhang2024clay,wu2024direct3d} for its simplicity and efficiency.
As shown in Fig.~\ref{fig:vae} (a), given an input garment mesh, we first uniformly sample a set of points (denoted as $X\in \mathrm{R}^{10000\times 3}$).
Next, we encode the sampled points to a set of vectors (denoted as $Z\in \mathrm{R}^{512\times 16}$) by applying cross-attention layers to $X$ and its downsampled version.
To effectively reconstruct garment meshes from latent vectors, we adopt the Unsigned Distance Field (UDF) representation~\cite{guillard2022udf}. The UDF is designed for non-watertight meshes and uses the gradient at each point to determine the presence of mesh surface.
Specifically, we feed query points ($\{q_{xyz}\}$) uniformly sampled from a 3D grid and the latent code $Z$ to few cross-attention layers to predict the Unsigned Distance for all $\{q_{xyz}\}$, which forms the UDF for the garment.
The garment mesh can then be extracted from the predicted UDF using MeshUDF~\cite{guillard2022udf}.
To train the VAE model, we follow prior works~\cite{de2023drapenet,Chen2024NeuralABC,guillard2022udf} and apply: a) a binary cross-entropy loss ($L_{bce}$) to the predicted Unsigned Distance, b) the L2 loss ($L_{grad}$) to the gradient at each query point and c) the KL divergence loss ($L_{KL}$) to the latent code $Z$. The full loss for the VAE training is $L = L_{bce} + \lambda_{grad}L_{grad} + \lambda_{KL}L_{KL}$, where $\lambda_{grad}, \lambda_{KL}$ are empirically set to $0.0001, 0.1$ in our experiments.

\paragraph{Text-based garment diffusion model.}
Given the learned vectorized latent space, we train a text-conditioned latent diffusion model for garment mesh generation.
As shown in Fig.~\ref{fig:vae}(b), for a latent vector $Z$, we add random noise $\epsilon$ sampled from a Normal Distribution to $Z$ and adopt a transformer-based network (denoted as $\mathcal{D}$) to recover the added noise.
To inject text prompts as conditions, we use the BERT model~\cite{devlin2018bert} to extract a text embedding $e\in \mathrm{R}^{768}$ and fuse it with latent features through cross-attention layers.
The objective to train the diffusion model follows the EDM~\cite{Karras2022edm,zhang20233dshape2vecset} and can be formulated as:
\begin{equation}
    L(\theta_{\mathcal{D}}) = \mathbb{E}_{\epsilon\in \mathcal{N}(0,\sigma^2\mathbf{I})}\|\mathcal{D}(Z+\epsilon, \sigma, e)) - Z\|_{2}^{2}
    \vspace{-2mm}
\end{equation}
where $\theta_{\mathcal{D}}$ denotes the trainable parameters of the denoiser network $\mathcal{D}$ and $\sigma$ is a randomly chosen noise level.
During inference, a randomly sampled noise vector is gradually denoised to a latent code $Z$ matching the given text prompt, which is then decoded to a garment mesh using the decoder in the VAE model introduced earlier.

\subsection{Gaussian Avatar Optimization}
\vspace{-1mm}
To further obtain reasonable geometry of hair strands and body shape, we utilize HAAR~\cite{sklyarova2024haar} which is a text-conditioned diffusion model, and the BodyShapeGPT~\cite{arbol2024bodyshapegpt} which is a GPT based LLM model to predict hair strands and SMPL shape parameters from the given text.
Given the garment mesh, hair strands, and SMPL body mesh, we attach 3D Gaussians to each component and optimize the attributes of these Gaussians using foundational diffusion models.
We choose 3D Gaussians for texture modeling for two reasons. First, compared to implicit representations such as SDF, 3D Gaussians are more flexible for animation and can be easily driven by meshes~\cite{yuan2024gavatar,qian2023gaussianavatars}. Second, 3D Gaussians have shown impressive capacity to capture high-fidelity appearance, especially when combined with diffusion models priors~\cite{yuan2024gavatar,liu2023humangaussian}.
Next, we discuss how to customize 3D Gaussians for each body part to ensure they maximally conform to the structure and motion of that part. 

\label{sec:gaussian_avatar}
\paragraph{3D Gaussians on body and garment meshes.} We use a similar strategy to combine 3D Gaussians with the body and garment part since their coarse geometry is both represented as 3D meshes.
For the body mesh, we utilize a SMPL~\cite{SMPL:2015} mesh $\Omega = \text{LBS}(\theta,\beta)$, where $\theta$ and $\beta$ are the SMPL pose and shape parameters, and LBS is the linear blend skinning function.
The garment mesh is generated and simulated as discussed in Sec.~\ref{sec:diffusion} and Sec.~\ref{sec:preliminaries}.
Below we discuss how to attach and deform a 3D Gaussian to a mesh $\mathcal{M}$, which could be a body or garment mesh.
Specifically, we associate each Gaussian $G_i{=}\{\mu_i, r_i, s_i, f_i, o_i\}$ with a face of the mesh $\mathcal{M}$ and define its position $\mu_i \in \mathbb{R}^3$, rotation $r_i \in \mathbb{R}^3$, and scaling $s_i \in \mathbb{R}^3$ in the face's local coordinate, as well as its color features $f_i\in \mathbb{R}^{d_c} $ and opacity $o_i$, where $d_c$ is the dimension of the Spherical Harmonic coefficients. 
The face's coordinate $\{P(\theta), R(\theta), k\}$ is defined similarly to \cite{qian2023gaussianavatars}, where the origin $P(\theta) \in \mathbb{R}^3$ is computed as the mean position of the face vertices, and the rotation matrix $R(\theta) \in \mathbb{R}^{3\times3}$ is formed by concatenating one edge vector of the face, the normal vector, and their cross product. We also compute a scalar $k$ by the mean length of the edges. 
The global Gaussian position, rotation, and scale $\{\hat \mu_i, \hat r_i, \hat s_i\}$ can be computed by applying the local-to-global transform:
\begin{equation}
\vspace{-1mm}
\label{eq:gaussian1}
    \begin{aligned}
    \hat{\mu}_i(\theta) &= k R(\theta)\cdot p_i + P(\theta)\\
    \hat{r}_i(\theta) &= R(\theta) \cdot r_i \\
    \hat{s}_i(\theta) &= k s_i
\end{aligned}
\end{equation}
We initialize the 3D Gaussians by uniformly sampling points on the mesh surface, the face correspondences are maintained throughout the Gaussian densification process~\cite{kerbl3Dgaussians}.
As observed in \cite{yuan2024gavatar}, directly optimizing Gaussian attributes with high-variance diffusion prior-based objective leads to noisy avatars. Thus, we adopt an implicit field $\mathcal{F}_\phi$ with parameters $\phi$ to model the Gaussian attributes. Specifically, we query the color features $f_i$, opacity $o_i$ of each Gaussian using its global position $\hat{p}_i(\tilde{\theta})$ under a canonical pose $\tilde{\theta}$ by $(f_i, o_i) = \mathcal{F}_\phi(\hat{\mu}_i(\tilde{\theta}))\,$. 
Note that we learn two separate implicit fields for the body $\mathcal{F}_\phi^b$ and garment $\mathcal{F}_\phi^g$ to prevent texture entanglement. The canonical garment mesh is a garment draped on the SMPL body in T-pose.
As a result, the 3D Gaussians attached to the body or garment mesh can be smoothly driven by the avatar's motion by Eq.~\ref{eq:gaussian1}.

\input{figs/qualitative_static}

\paragraph{3D Gaussians on hair strands.}
Different from the body and garment meshes, hair strands have a more delicate and complex structure.
As a result, the Gaussians associated with the strands are expected to have a long and thin shape. 
Inspired by existing works that adopt 3D Gaussians for hair reconstruction~\cite{zakharov2024gh,luo2024gaussianhair}, we assign one 3D Gaussian to each line segment in each hair strand. 
As shown in the green circle in Fig.~\ref{fig:overview}, for a line segment defined by two endpoints $l_i$ and $l_{i+1}$, we assign a Gaussian denoted as $G_{i}=\{\mu_i, r_i, s_i, f_i, o_i\}$ to model the segment. To enforce the Gaussians to be consistent with the hair strands, we explicitly compute the position, rotation, and scale of $G_i$ by:
\begin{equation}
\vspace{-1mm}
\begin{aligned}
\label{eq:hair}
    \mu_i &= (l_i + l_{i+1}) / 2 \\
    s_i &= [\norm{l_{i+1}-l_i} / 2, \gamma, \gamma] \\
    r_i &= [1 + \mu_i\cdot d_i, u\times d_i], 
\end{aligned}
\vspace{-1mm}
\end{equation}
where $d_i$ is the direction of a hair strand, and $u$ are unit vectors along the $x$-axis. We set $\gamma=0.001$ to keep the hair Gaussians thin. To learn opacity $\{o_i\}$ and color $\{f_i\}$ for the hair Gaussians, we similarly train an implicit field $\mathcal{F}^h_\phi$ as discussed in the garment and body part above. During inference, the hair Gaussians can be driven by the hair strands smoothly and consistently using Eq.~\ref{eq:hair}.

\paragraph{Shading Model.} Lighting plays an important role in modeling appearance details in motion such as garment wrinkles. To encourage the 3D Gaussians to capture pose-independent albedo without baked-in shading, we incorporate a Phong shading model~\cite{wang2023disentangled} into our pipeline. 
Since the normal for each Gaussian is noisy, we instead use the normal of its corresponding face (denoted as $n_p$) in the lighting model. 
To mimic random lighting, we sample the point light position $l_p\in \mathbb{R}^3$, color $l_c\in \mathbb{R}^3$, as well as an ambient light color $l_a\in \mathbb{R}^3$. The shaded color of each 3D Gaussian can then be computed by ${f_i}'=f_i\cdot(\text{max}(0,n_p\cdot(l_p - \hat \mu_i)/\norm{l_p - \hat \mu_i})\cdot l_c + l_a)$, where $\hat{\mu}_i$ is the global coordinate of the Gaussian computed by Eq.~\ref{eq:gaussian1}. 
%

\paragraph{Gaussian avatar optimization.} To learn the implicit fields for the hair, body, and garment parts, we utilize the Score Distillation Sampling (SDS)~\cite{poole2022dreamfusion} objective with a pre-trained text-to-image diffusion model~\cite{rv5.1} to supervise the Gaussian splatting-based rendering $I(\eta)$ of our model, where $\eta$ denotes all learnable parameters \ie, $\eta=\{\mathcal{F}_\phi^b, \mathcal{F}_\phi^g, \mathcal{F}_\phi^h\}$. 
Given a text prompt $\mathcal{T}$ and the noise prediction $\hat{\epsilon}(I_t; \mathcal{T}, t)$ of the diffusion model, SDS optimizes parameters $\eta$ by minimizing the difference between the noise $\epsilon$ added to the rendered image $I(\eta)$ and the predicted noise $\hat{\epsilon}$ by the diffusion model:
\begin{equation}
\label{eq:sds}
    \nabla_{\eta} \mathcal{L}_\text{SDS} = E_{t,\epsilon} \left[w(t)(\hat{\epsilon}(I_t; \mathcal{T}, t) - \epsilon) \frac{\partial I}{\partial \eta} \right]\,,
\vspace{-2mm}
\end{equation}
where $t$ is the noise level, $I_t$ is the noised image, and $w(t)$ is a weighting function. 
Additionally, to prevent broken hairs caused by transparent Gaussians in the middle of a hair strand, we propose a regularization on the opacity of hair Gaussians.
Our key intuition is that along a hair strand, the Gaussians closer to the scalp (e.g., hair roots) should have larger opacity values than the Gaussians further away from the scalp (e.g., hair ends).
This way the optimization process is able to trim the extra hair to match the prompt while avoiding the broken hair artifacts.
Specifically, given a hair point cloud $h_0\in \mathrm{R}^{N_s\times N_l\times 3}$ as discussed in Sec.~\ref{sec:preliminaries}, we sample the opacity ($o\in \mathrm{R}^{N_s\times N_l}$) of its associated Gaussians from the learned implicit fields (i.e., $\mathcal{F}_\phi^h$ discussed in Sec.~\ref{sec:gaussian_avatar}) and add the following regularization:
\begin{equation}
    L_{hair} = \frac{1}{N_sN_l}\sum_{i=1}^{N_s}{\sum_{j=2}^{N_l}{(o_{i,j-1}-o_{i,j})}}
\vspace{-2mm}
\end{equation}
Our final objective is $L = L_{SDS} + \lambda_{hair}L_{hair}$, where $\lambda_{hair}$ is empirically set to $1.0$ in our experiments.

%% file: figs/overview.tex
\begin{figure*}[t]
    \centering    
    \includegraphics[width=0.98\textwidth]{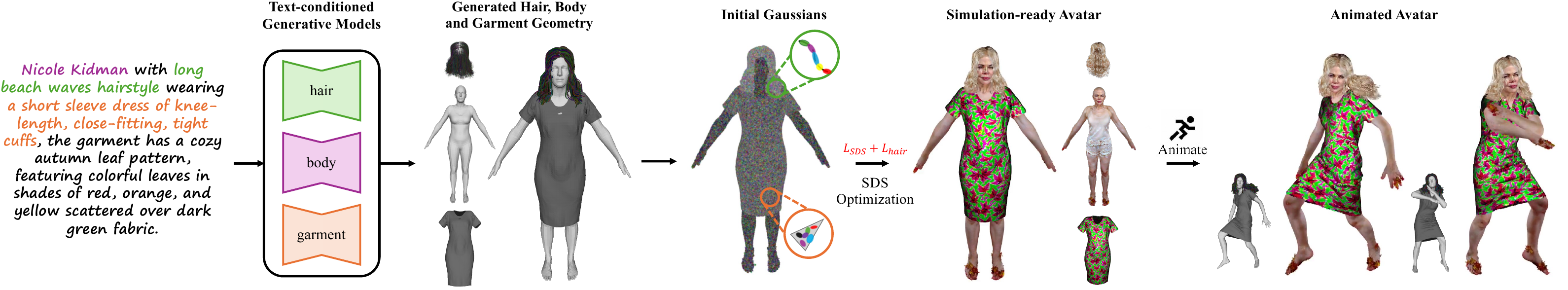}
    \caption{\textbf{Overview.} Given a text prompt, \acronym first generates hair strands, body mesh, and garment using their respective text-conditioned generative models. We then bring them together using physics simulation, and attach 3D Gaussians to learn their appearances and to adapt them according to the text prompts.  We assign one 3D Gaussian to each line segment in the hair strands with their length significantly larger than their diameter (green circle), whereas the Gaussians for meshes are defined within a local coordinate system of each face (orange circle). We optimize the properties of all 3D Gaussians through the Score Distillation Sampling (SDS) loss using image-based diffusion models, and a novel regularization $L_{hair}$ for hairs to ensure plausible hair structure. The generated 3D avatar can be simulated by any pose sequence showing realistic and dynamic motion effects such as flowing hairs and garment wrinkles.
}
    \label{fig:overview}
\end{figure*}

%% file: figs/vae.tex
\begin{figure*}[t]
    \centering    
    \includegraphics[trim={0 0 0 0.5cm},clip,width=0.98\textwidth]{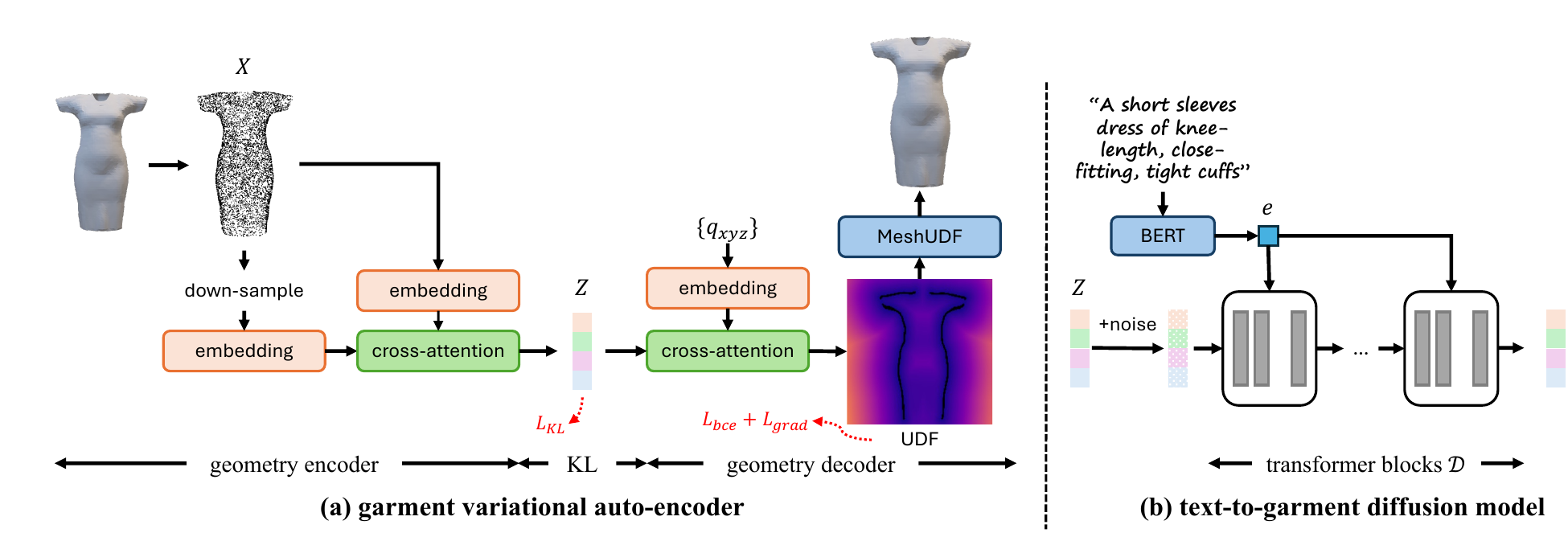}
    \vspace{-2mm}
    \caption{\textbf{Text-based garment diffusion model.} See Sec.~\ref{sec:diffusion} for details.
}
    \label{fig:vae} 
\end{figure*}

%% file: figs/qualitative_static.tex
\begin{figure*}[t]
    \centering    
    \includegraphics[width=0.9\textwidth]{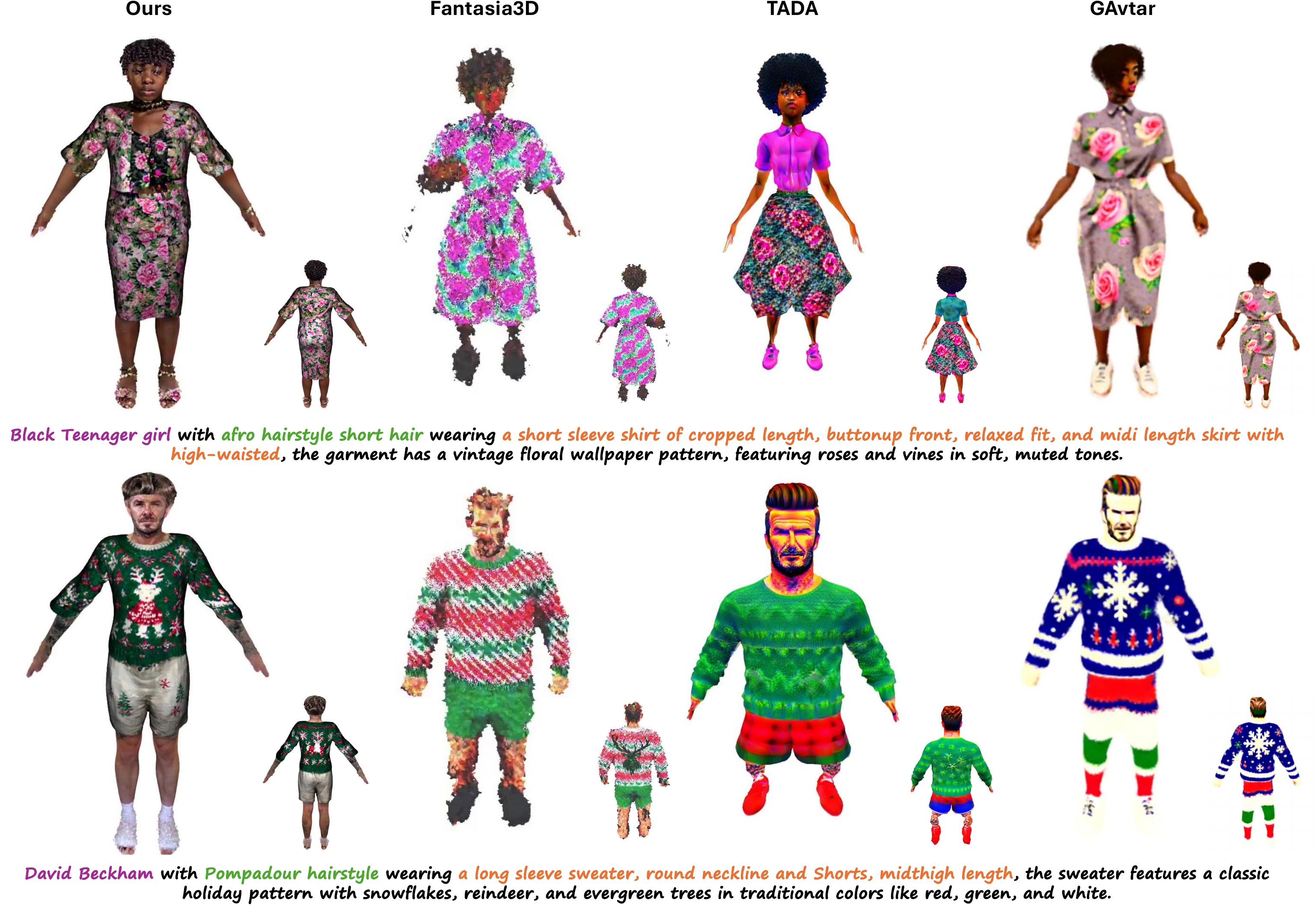}
    \vspace{-3mm}
    \caption{\textbf{Qualitative comparison with the state-of-the-art methods.} See the supplementary video for more comparisons.
}
    \label{fig:qulitative_static}
\end{figure*}

%% file: sec/4.Experiments.tex
\section{Experiments}
\vspace{-2mm}
\input{figs/qualitative_dynamic}

\subsection{Datasets.}
\vspace{-1mm}
To learn the text-based garment diffusion model discussed in Sec.~\ref{sec:diffusion}, we use the Garment Pattern Generator (GPG) dataset~\cite{korosteleva2021generating} and the CLOTH3D~\cite{bertiche2020cloth3d} dataset processed by~\cite{de2023drapenet}. 
For text prompts, we leverage the prompt annotations for the GPG dataset from~\cite{he2024dresscode}. Since the CLOTH3D dataset does not include paired prompt annotation, we render each garment on top of the SMPL body mesh and query the GPT4v~\cite{achiam2023gpt} with predefined questions to generate a prompt for each garment mesh, describing its type, shape, length, and width.
Together we build a dataset containing around 20000 meshes with paired prompt annotations. The dataset covers common garment types such as t-shirts, tank tops, jackets, shorts, pants, skirts, dresses, etc. 


%
 %

\subsection{Qualitative Evaluations.}
\vspace{-1mm}
\label{sec:diffusion_eval}

We compare our synthesized avatars with start-of-the-art text-driven 3D avatar generation methods, including Fantasia3D~\cite{chen2023fantasia3d}, TADA~\cite{liao2024tada} and GAvatar~\cite{yuan2024gavatar}.
First, we present qualitative comparisons of synthesized avatars in a canonical pose in Fig.~\ref{fig:qulitative_static}. Our method not only captures diverse garment geometry (e.g., skirts, tops, and bottoms) but also produces superior texture compared to baseline models.
Next, we demonstrate animated avatars in Fig.~\ref{fig:qulitative_dynamic}. Note that since Fantasia3D utilizes implicit representations, it cannot be readily animated. 
GAvatar and TADA model the human body and garment by a single geometry representation. For animation, they attach garments to the human body and use linear blending skinning for deformation. As a result, they fail to produce realistic motion for loose garments such as dresses. As shown by the first row of Fig.~\ref{fig:qulitative_dynamic}, when the avatar raises her legs, only our model produces physically plausible motion for the dresses, whereas both baselines undesirably separate the dresses into two pieces. We provide abundant animation comparisons in the supplementary video to showcase the difference.

\subsection{Quantitative Evaluations.}
\vspace{-1mm}

\begin{table}[h]
\centering
\footnotesize
\begin{tabular}{ccccc}
\hline
Metrics & TADA & Fantasia3D & GAvatar \\
\hline
Appearance preference & 89.55 & 100 & 87.03 \\
\hline
Motion preference & 91.87 & 100 & 94.47 \\
\hline
\end{tabular}
\vspace{-1mm}
\caption{\textbf{User study.} We provide the preference percentage of our method against state-of-the-art methods. The higher score indicates that more users favor our method. \acronym significantly outperforms other methods in terms of user preference. We encourage looking at the supplementary video.}
  \label{table::user_study}
  \vspace{-0.15in}
\end{table}

\paragraph{User study.}
We follow existing works~\cite{yuan2024gavatar, kolotouros2023dreamhuman, liao2024tada} and quantitatively evaluate \acronym through an extensive A/B user study. 
Specifically, we run \acronym and baseline methods on 22 prompts. For comparison, we generate a static 360 video of the avatar in A-pose and a video showing the avatar driven by a motion sequence. 
At each round, we present a pair of results by our method and one of the baselines to a user. 
For the static 360 video pairs, we ask the user to choose the method with a better appearance. For the animated video pairs, we ask the user to pay more attention to the motion. We collected 540 votes from 18 users and present the results in Table.~\ref{table::user_study}. For both appearance and motion, \acronym is consistently favored by users, demonstrating its superior performance compared to baselines.



%% file: figs/qualitative_dynamic.tex
\begin{figure*}[h]
    \centering    
    \includegraphics[width=0.9\textwidth]{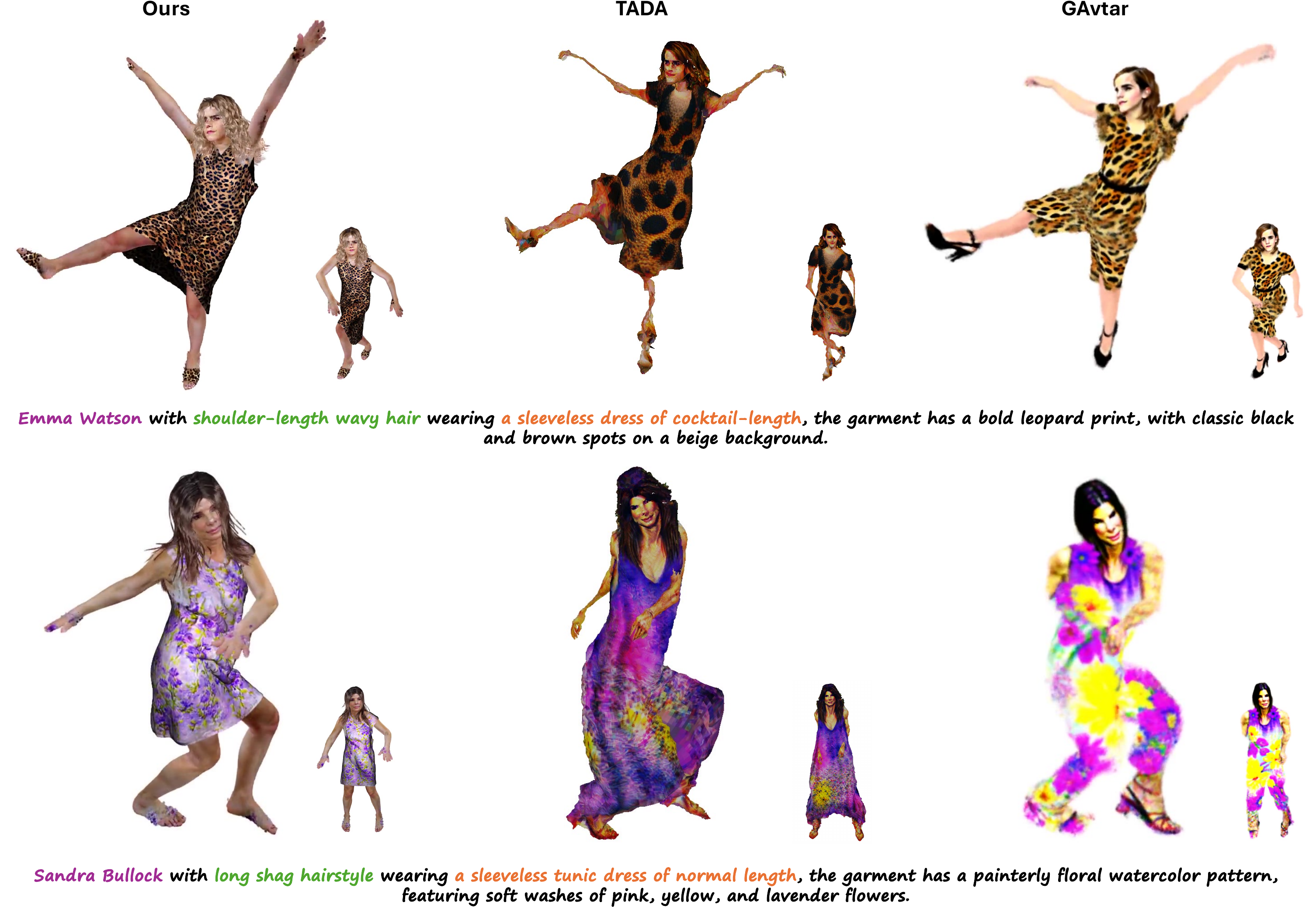}
    \vspace{-3mm}
    \caption{\textbf{Qualitative comparison of animated avatars.} See the supplementary video for more animation comparisons. \vspace{-5mm}
}
    \label{fig:qulitative_dynamic}
\end{figure*}

%% file: sec/5.Conclusions.tex
\section{Conclusions and Future Work}
\vspace{-2mm}
In this work, we proposed a novel framework for generating simulation-ready avatars with layered hair and clothing from a text prompt. Unlike existing methods for 3D human avatar generation, our approach represents the human body, hair, and garments in a layered and simulation-ready structure. 
To this end, we first generated garment mesh, hair strands, and body shape parameters from text-conditioned generative models.  We then optimized 3D Gaussians attached to each region to learn realistic appearance. The generated avatars can be seamlessly animated using physics simulators. 
Both qualitative and quantitative results demonstrate the effectiveness of our method in generating realistic, animatable clothed avatars, allowing cloth, and hair simulation with dynamic effects such as wrinkles and flowing hair, which surpasses state-of-the-art techniques. 

While SimAvatar provides avatars with significantly enhanced realism, there are still areas for further improvements. Currently, our hair and garment generation models are trained on specific datasets, which may bottleneck the diversity due to the used training data. In future work, we aim to explore methods for generating simulation-ready avatars that extend beyond the existing dataset diversity. Additionally, SimAvatar simulates garments and hair sequentially, which can fail in certain cases, such as avatars wearing hoods. Implementing joint simulation for hair and garments would allow us to accommodate more complex garments. Finally, accessories and footwear remain entangled with the body or garment layers; in the future, we plan to focus on generating fully disentangled avatars.

%% file: supplementary.tex
\appendix

\section{Quantitative Evaluations}
\vspace{-1mm}
\label{sec:vqa}
\paragraph{VQAScore}
We quantitatively compare our method against baseline methods using the VQAScore~\cite{lin2024evaluating}. Instead of treating the text prompt as a set of unordered words, the VQAScore assesses alignment between prompts and generated assets by posing targeted questions to foundational models. This enables evaluation of more complex prompts involving multiple entities and relationships, providing results that align more closely with human perceptual evaluation~\cite{lin2024evaluating} compared to the CLIP score. The VQAScore is particularly well-suited to our task, which involves generating avatars with compositional text prompts specifying different parts (e.g., hair, identity, and garments). As shown in Table~\ref{table::vqa}, our model achieves a significantly higher VQAScore than baseline methods, demonstrating its superior ability to align with input prompts.

\paragraph{CLIP-Score} Following prior work, we report the CLIP score of our method and baselines in Table~\ref{table::vqa}. However, it is important to note that the CLIP score has been widely observed to be unreliable for accurately assessing visual quality and alignment with text prompts~\cite{lin2024evaluating}. In Fig.~\ref{fig:qualitative_comparison_1}, we additionally present both the VQA-Score (green) and the CLIP score (red) for each avatar. Overall, the VQA-Score demonstrates a closer alignment with human perception. That said, since visual quality is inherently subjective, we encourage readers to examine the figures and accompanying videos for a more holistic evaluation.

\begin{table}[h]
\centering
\footnotesize
\begin{tabular}{cccccc}
\hline
Metrics & TADA & Fantasia3D & GAvatar & HG & Ours \\
\hline
VQA score $\uparrow$ & 0.45 & 0.38 & 0.44 & 0.53 & \textbf{0.75} \\
\hline
CLIP $\uparrow$ & 33.50 & 37.44 & 30.74 & 30.81 & 33.39 \\
\hline
\end{tabular}
\vspace{-2mm}
\caption{\textbf{Quantitative Evaluations.} We quantitatively compare the VQAScore~\cite{lin2024evaluating} and CLIP-score with TADA~\cite{liao2024tada}, Fantasia3D~\cite{chen2023fantasia3d}, GAvatar~\cite{yuan2024gavatar} and HumanGaussians (HG)~\cite{liu2023humangaussian}.}
  \label{table::vqa}
  \vspace{-0.15in}
\end{table}

\section{Implementation Details}
\vspace{-1mm}
\label{sec:implementation}

\paragraph{Text-to-garment diffusion model training.} To generate garment meshes from text prompts, we train the VAE and diffusion model sequentially, as described in Sec~\ref{sec:diffusion} of the main paper. The VAE is trained with a batch size of 14, while the diffusion model uses a batch size of 24. The entire training process takes approximately one week on four V100 GPUs.

\paragraph{Gaussian avatar learning.} We uniformly sample $2.6 \times 10^5$ and $10^6$ Gaussians on the body and garment mesh surface as the initial Gaussian positions. 
During optimization, we combine the rendered avatar image with a solid background of random color.
We optimize the implicit fields (i.e., $\{\mathcal{F}_\phi^b, \mathcal{F}_\phi^g, \mathcal{F}_\phi^h\}$) with a learning rate of 0.001. The optimization takes around six hours per avatar on a single V100 GPU.
To facilitate stable training, we first optimize the Gaussians attached to inner layer (i.e., hair and body) for 4000 iterations. We then include the garment layer and optimize the avatar for another 6000 iterations. 

\section{Layer-wise Training Strategy}
\vspace{-1mm}
\label{sec:prompt}
To facilitate disentanglement of the hair, body and garment, we separately render each layer and pair them with different prompts.
For example, given the prompt: ``Adele with long layered waves hairstyle wearing a sleeveless dress of  tea-length, gathered waist, the garment has a delicate butterfly print, with small, colorful butterflies scattered across the neckline and sleeves.'', we render the hair, body, and garment layers individually. The corresponding prompts used are: "long layered waves hairstyle" for the hair, "Adele in a tank top and shorts" for the body, and "a sleeveless tea-length dress with a gathered waist, featuring a delicate butterfly print with small, colorful butterflies scattered across the neckline and sleeves" for the garment.
To further disentangle face and hair (e.g., preventing hair textures from appearing in the face region), we render the avatar's head in a zoomed-in view and pair it with the prompt: "Adele with short buzz hair and a bold forehead." This approach significantly reduces the likelihood of hair textures being learned by the body region, as will be demonstrated in Sec.~\ref{sec:ablation}.
In addition to rendering views of the face, body, hair, and garment, we also zoom in on specific parts such as the hands, feet, lower body, and upper body to enhance the quality of the avatar's details in these regions.

\section{Ablation Studies}
\vspace{-1mm}
In this section, we evaluate the effectiveness of different modules and present qualitative results in Fig.~\ref{fig:ablation}.
\label{sec:ablation}
\input{figs/supp/ablation}
\paragraph{Hair constraint.}
To address the issue of broken hairs during Gaussian avatar optimization, we introduced a hair constraint, as described in Sec.~\ref{sec:gaussian_avatar} of the main paper. As illustrated in Fig.~\ref{fig:ablation}(a)(b), avatars trained without the hair constraint exhibit broken hairs and floating Gaussians around the head. These results highlight the effectiveness of the hair constraint in producing cohesive and realistic hair representations.

\paragraph{Prompt engineering.}
To achieve complete disentanglement of hair, body, and garment, we propose optimizing each layer separately while pairing the optimization with distinct prompts. Fig.~\ref{fig:ablation}(c)(d) compares the results of using the prompt "buzz cut, bold forehead" for face-view optimization versus not using it. Without this prompt, the model generates hair textures on the body, hindering the full disentanglement of the different parts.

\paragraph{Disentangled training strategy.} As outlined above, during training, we optimize each layer (i.e., hair, body, and garment) separately using distinct prompts. To validate the effectiveness of this training strategy, we conducted an experiment where only the full avatar was rendered and optimized by the SDS loss without layer separation. As shown in Fig.~\ref{fig:ablation}(e)(f), this approach fails to produce a coherent avatar and results in entanglement between body and garments (e.g., garment textures appearing on the body).

\section{More Qualitative Results}
\vspace{-1mm}
\label{sec:qualitative_supp}
\input{figs/supp/quali1_supp}
\input{figs/supp/quali2_supp}
\input{figs/supp/layers1}
\input{figs/supp/layers2}
We present comparisons with baselines in Fig.~\ref{fig:qualitative_comparison_1} and Fig.~\ref{fig:qualitative_comparison_2}. For a more comprehensive evaluation of motion, we strongly encourage readers to refer to the accompanying video.
Additionally, in Fig.~\ref{fig:qualitative_layers_1} and Fig~\ref{fig:qualitative_layers_2}, we provide further results of our method, illustrating the geometry and texture of each individual layer, including hair, face, garment, body, and the full avatar.

%% file: figs/supp/ablation.tex
\begin{figure}[h]
    \centering    
    \includegraphics[width=0.42\textwidth]{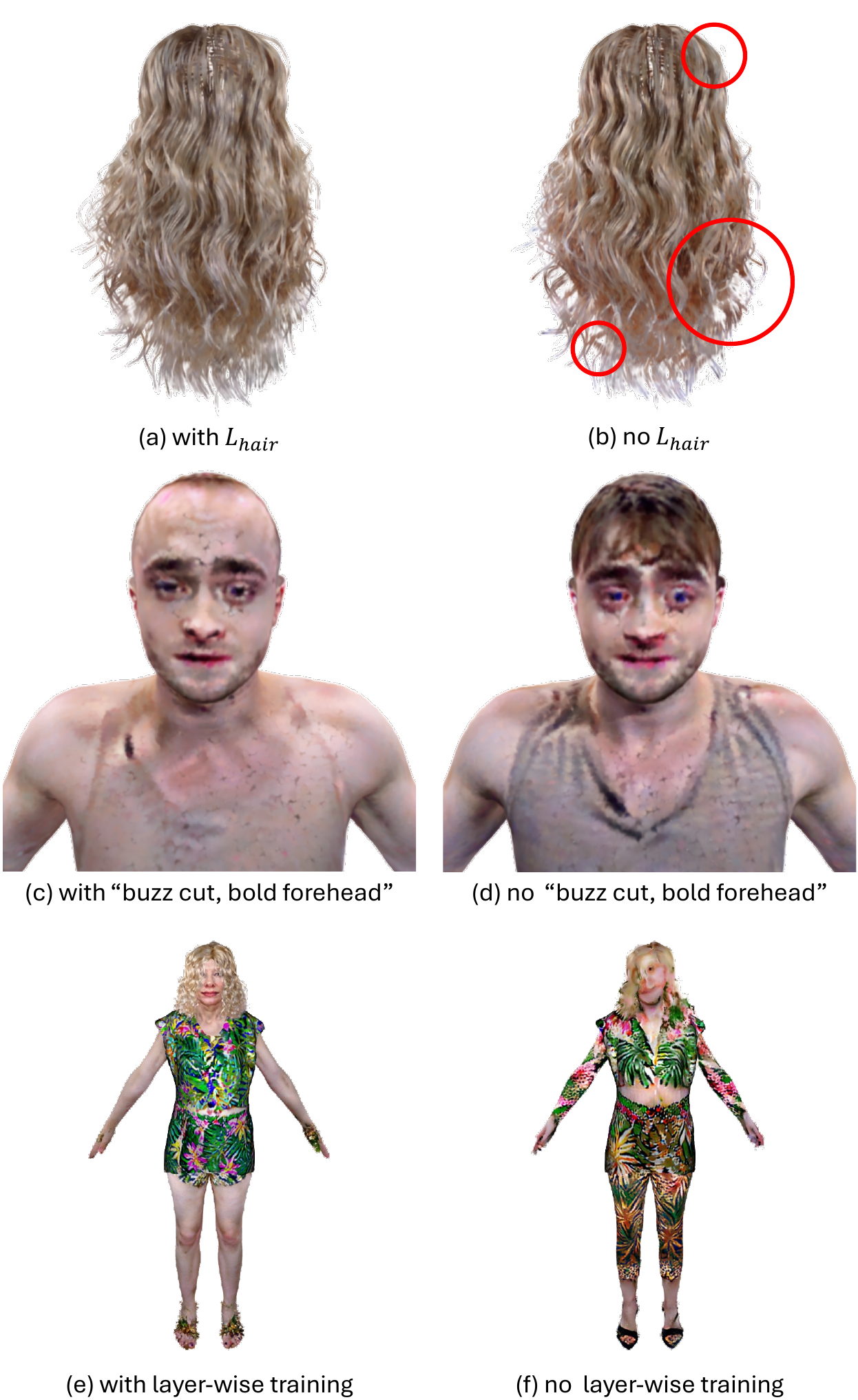}
    \vspace{-3mm}
    \caption{\textbf{Ablation Studies.} See Sec.~\ref{sec:ablation} for details. \vspace{-5mm}
}
    \label{fig:ablation}
\end{figure}

%% file: figs/supp/quali1_supp.tex
\begin{figure*}[h]
    \centering    
    \includegraphics[width=0.9\textwidth]{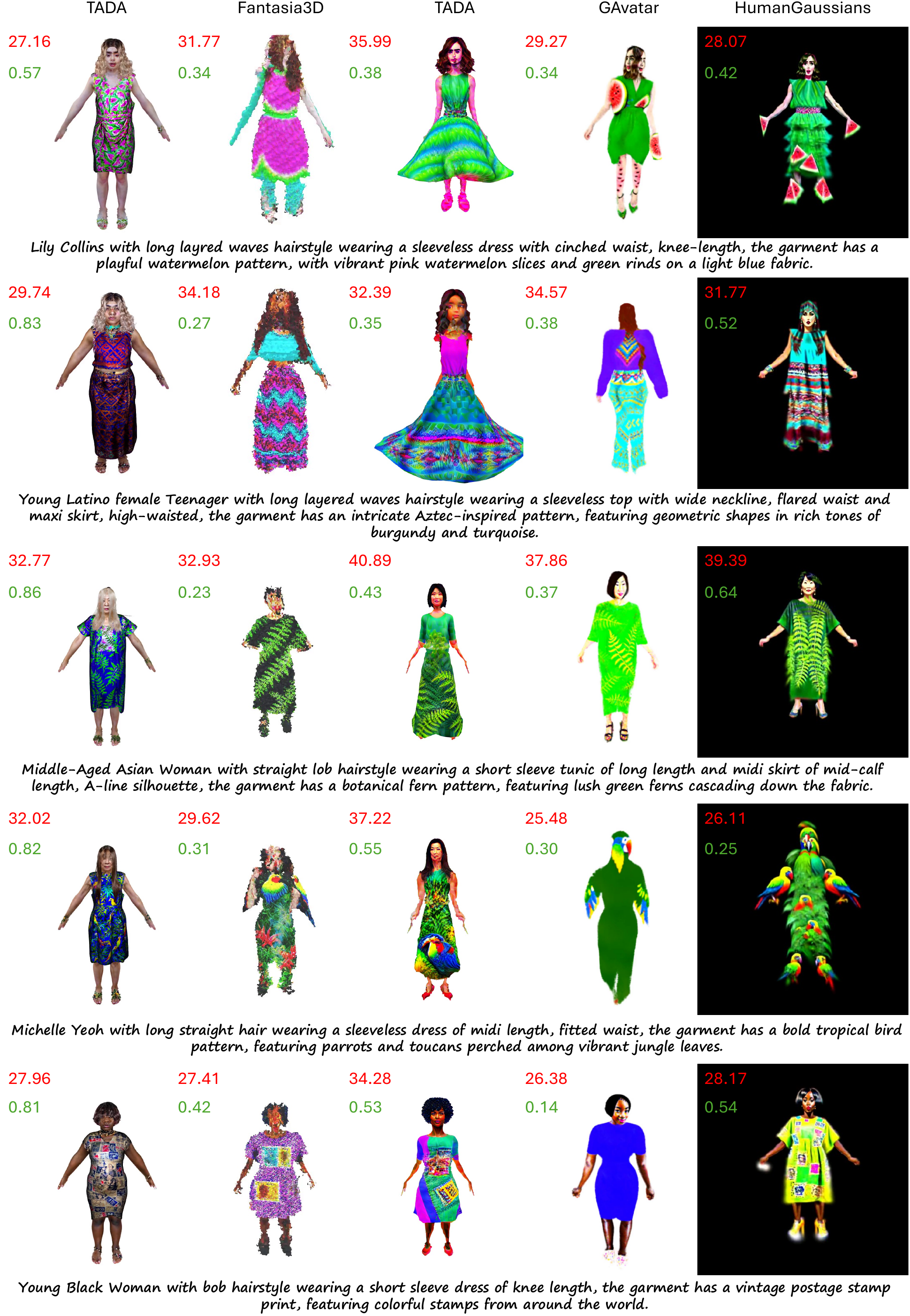}
    \vspace{-3mm}
    \caption{\textbf{More qualitative results.} Red and green numbers indicate CLIP and VQA score, respectively. \vspace{-5mm}
}
    \label{fig:qualitative_comparison_1}
\end{figure*}

%% file: figs/supp/quali2_supp.tex
\begin{figure*}[h]
    \centering    
    \includegraphics[width=0.9\textwidth]{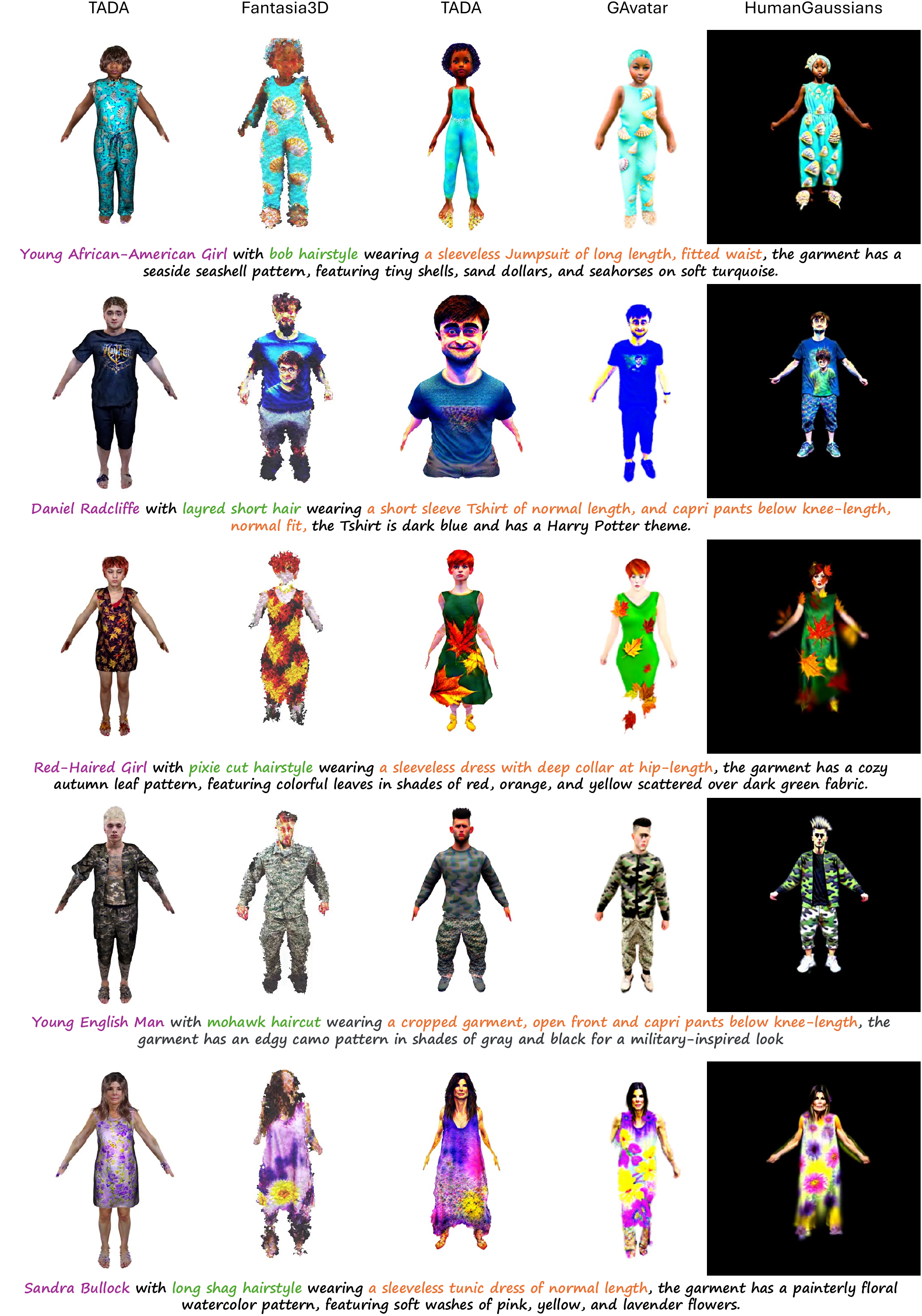}
    \vspace{-3mm}
    \caption{\textbf{More qualitative results.} \vspace{-5mm}
}
    \label{fig:qualitative_comparison_2}
\end{figure*}

%% file: figs/supp/layers1.tex
\begin{figure*}[h]
    \centering    
    \includegraphics[width=0.85\textwidth]{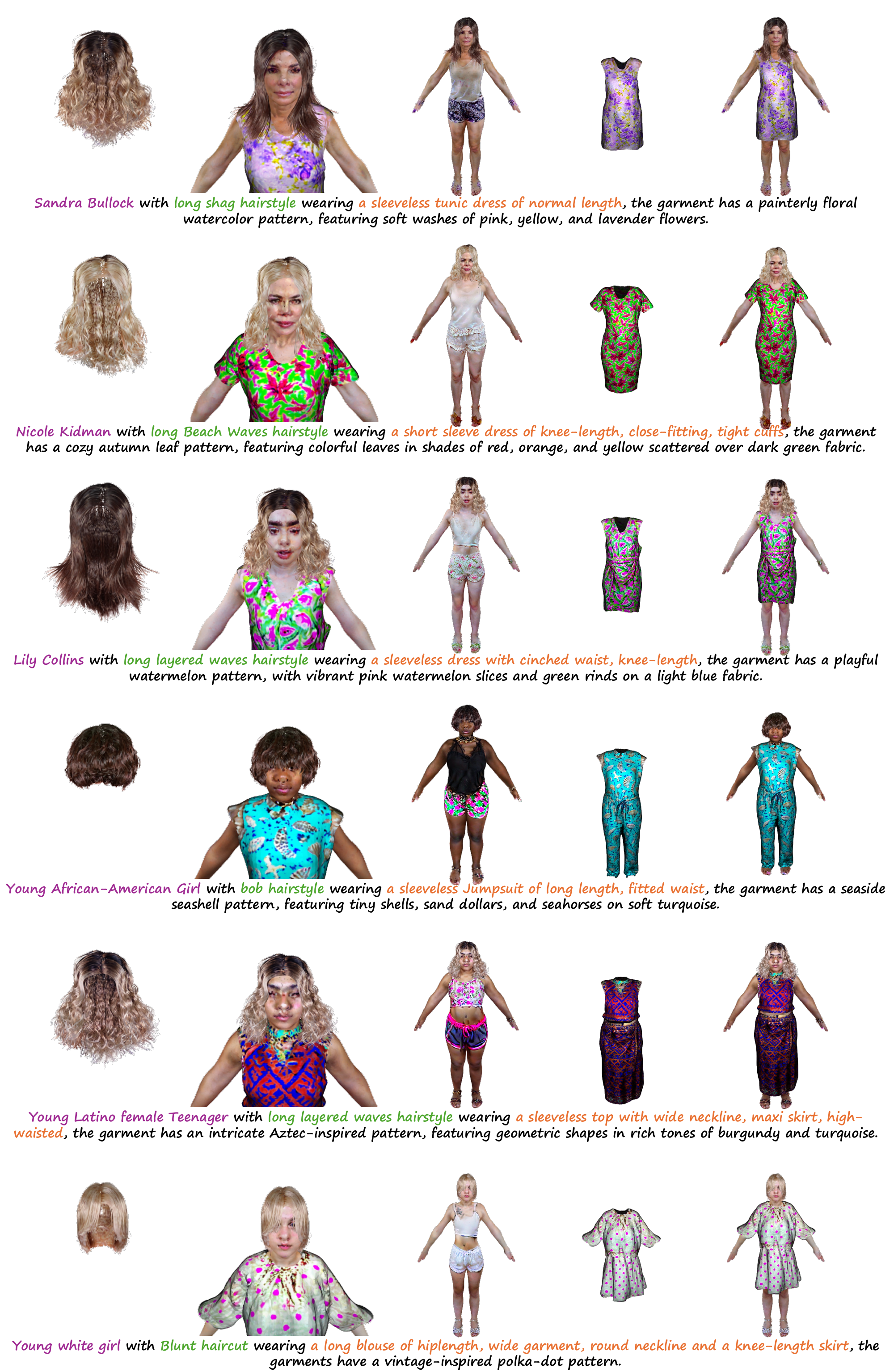}
    \vspace{-3mm}
    \caption{\textbf{Layer-wise visualization of SimAvatar.} 
}
    \label{fig:qualitative_layers_1}
\end{figure*}

%% file: figs/supp/layers2.tex
\begin{figure*}[h]
    \centering    
    \includegraphics[width=0.85\textwidth]{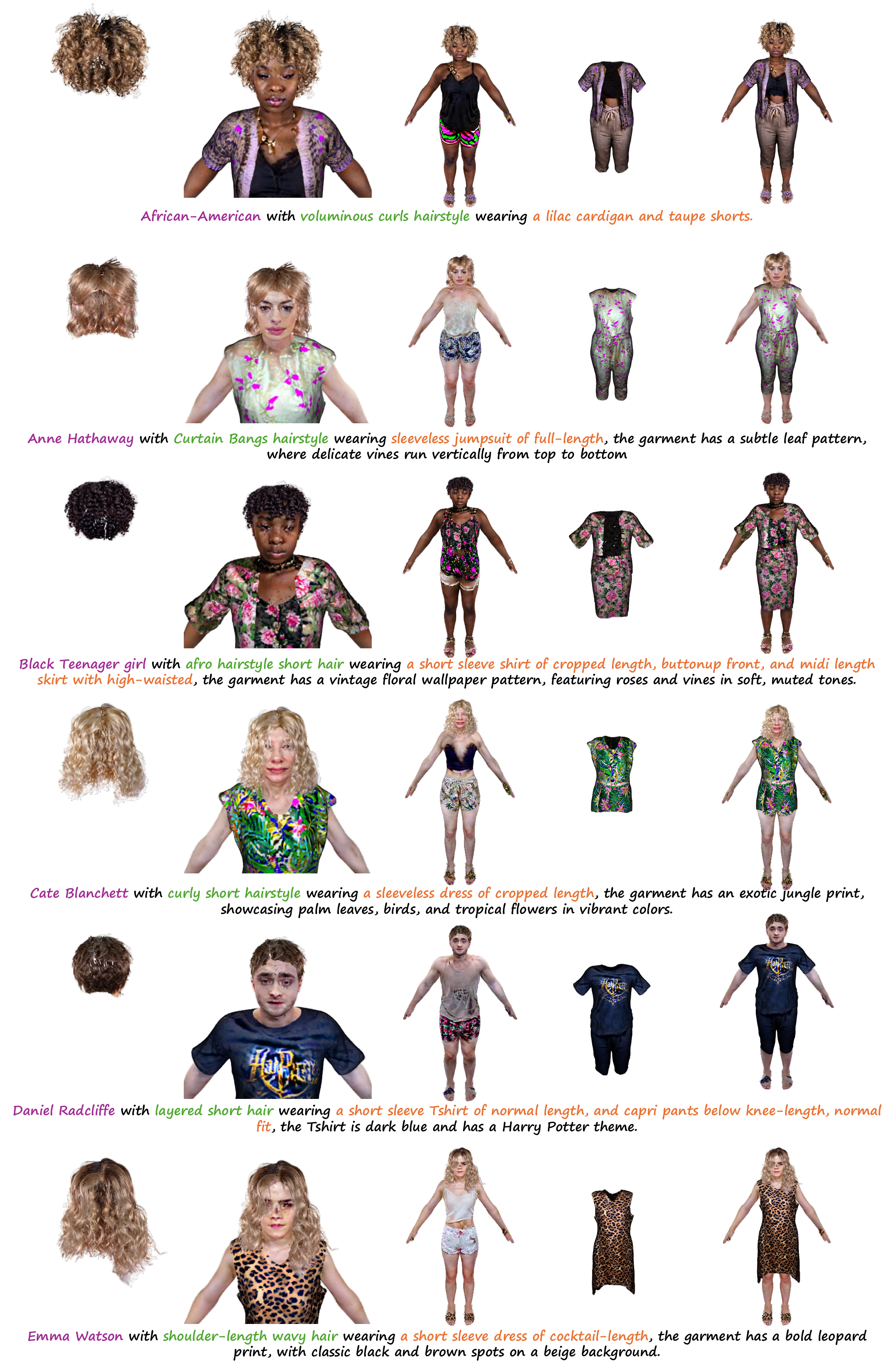}
    \vspace{-3mm}
    \caption{\textbf{Layer-wise visualization of SimAvatar.} 
}
    \label{fig:qualitative_layers_2}
\end{figure*}